\documentclass[conference]{IEEEtran}
\IEEEoverridecommandlockouts
\usepackage{cite}
\usepackage{amsmath,amssymb,amsfonts}
\usepackage{algorithmic}
\usepackage{graphicx}
\usepackage{textcomp}
\usepackage{xcolor}
\usepackage{multirow}
\def\BibTeX{{\rm B\kern-.05em{\sc i\kern-.025em b}\kern-.08em
    T\kern-.1667em\lower.7ex\hbox{E}\kern-.125emX}}
\begin{document}
	
\title{SR-LIO: LiDAR-Inertial Odometry with Sweep Reconstruction 
	\thanks{This work was not supported by any organization.}
}

\author{Zikang~Yuan$^{1}$, Fengtian~Lang$^{2}$, Tianle~Xu$^{2}$ and Xin~Yang$^{2}$
	\thanks{$^{1}$Zikang~Yuan is with Institute of Artificial Intelligence, Huazhong University of Science and Technology, Wuhan, 430074, China. (E-mail: {\tt\small yzk2020@hust.edu.cn})}%
	\thanks{$^{2}$Fengtian~Lang, Tianle~Xu and Xin~Yang are with the Electronic Information and Communications, Huazhong University of Science and Technology, Wuhan, 430074, China. (E-mail: {\tt\small U201913666@hust.edu.cn; tianlexu@hust.edu.cn; xinyang2014@hust.edu.cn})}%
}

\maketitle

\begin{abstract}
This paper proposes a novel LiDAR-Inertial odometry (LIO), named SR-LIO, based on an iterated extended Kalman filter (iEKF) framework. We adapt the sweep reconstruction method, which segments and reconstructs raw input sweeps from spinning LiDAR to obtain reconstructed sweeps with higher frequency. We found that such method can effectively reduce the time interval for each iterated state update, improving the state estimation accuracy and enabling the usage of iEKF framework for fusing high-frequency IMU and low-frequency LiDAR. To prevent inaccurate trajectory caused by multiple distortion correction to a particular point, we further propose to perform distortion correction for each segment. Experimental results on four public datasets demonstrate that our SR-LIO outperforms all existing state-of-the-art methods on accuracy, and reducing the time interval of iterated state update via the proposed sweep reconstruction can improve the accuracy and frequency of estimated states. The source code of SR-LIO is publicly available for the development of the community.
\end{abstract}

\begin{IEEEkeywords}
state estimation, SLAM, LiDAR-inertial fusion
\end{IEEEkeywords}

\section{Introduction}

Three-dimension light detection and ranging (LiDAR) can directly capture accurate and dense scene structure information in a large range and thus has become one of the mainstream sensors in outdoor robots and autonomous driving fields. An odometry utilizing only 3D LiDAR \cite{zhang2014loam, zhang2017low, shan2018lego, behley2018efficient, wang2021f, deschaud2018imls, dellenbach2022ct} has the ability to estimate accurate pose in most scenarios, and transform the point clouds collected at different times to a unified coordinate system. However, there are still the two main problems in LiDAR-only odometry: 1) Most, if not all, existing LiDAR odometry rely on the Iterative Closest Point algorithm for pose estimation while an inaccurate initial motion value can largely increase the time consumption. 2) For scenes without rich geometric structure information, the commonly used point-to-plane ICP algorithm usually fails due to lack of sufficiently reliable constraints for pose estimation. Introducing Inertial Measurement Unit (IMU) as an additional sensor is a promising solution to address the above two problems with little memory and time consumption.

Existing LiDAR-Inertial odometry (LIO) systems \cite{zhang2014loam, zhang2017low, shan2018lego, li2021towards, ye2019tightly, qin2020lins, shan2020lio, xu2021fast, xu2022fast, chen2023direct} can be divided into two groups: loosely-coupled and tightly-coupled. The loose-coupled framework \cite{zhang2014loam, zhang2017low, shan2018lego} mainly uses IMU measurements to calibrate the motion distortion of LiDAR point clouds, and provide motion priors for ICP pose estimation. On the basis of loosely-coupling, the tightly-coupled framework also uses IMU measurements to provide motion constraints together with ICP, so as to achieve more accurate and robust state estimation. The tightly-coupled LIO framework can be mainly categorized into three types: iterated extended Kalman filter (iEKF), bundle adjustment (BA) and graph optimization. The performance of all these three types both depends on the time interval of integrating IMU measurements. Specifically, for BA and graph optimization framework, a long integration time will lead to a large accumulative error of pre-integration. For iEKF framework, a long integration time will reduce the frequency of state update. In general, the less accumulative error in the predicted state, the more accurate and robust result can be estimated. However, the time interval between two consecutive sweeps of spinning LiDAR is 100\,ms, which makes IMU pre-integration or the predicted state has accumulative error. If some key-sweeps are selected from a series of consecutive sweeps, the time interval between two key-sweeps is even longer, further increasing the accumulative error.

\begin{figure}
	\begin{center}
		\includegraphics[scale=0.39]{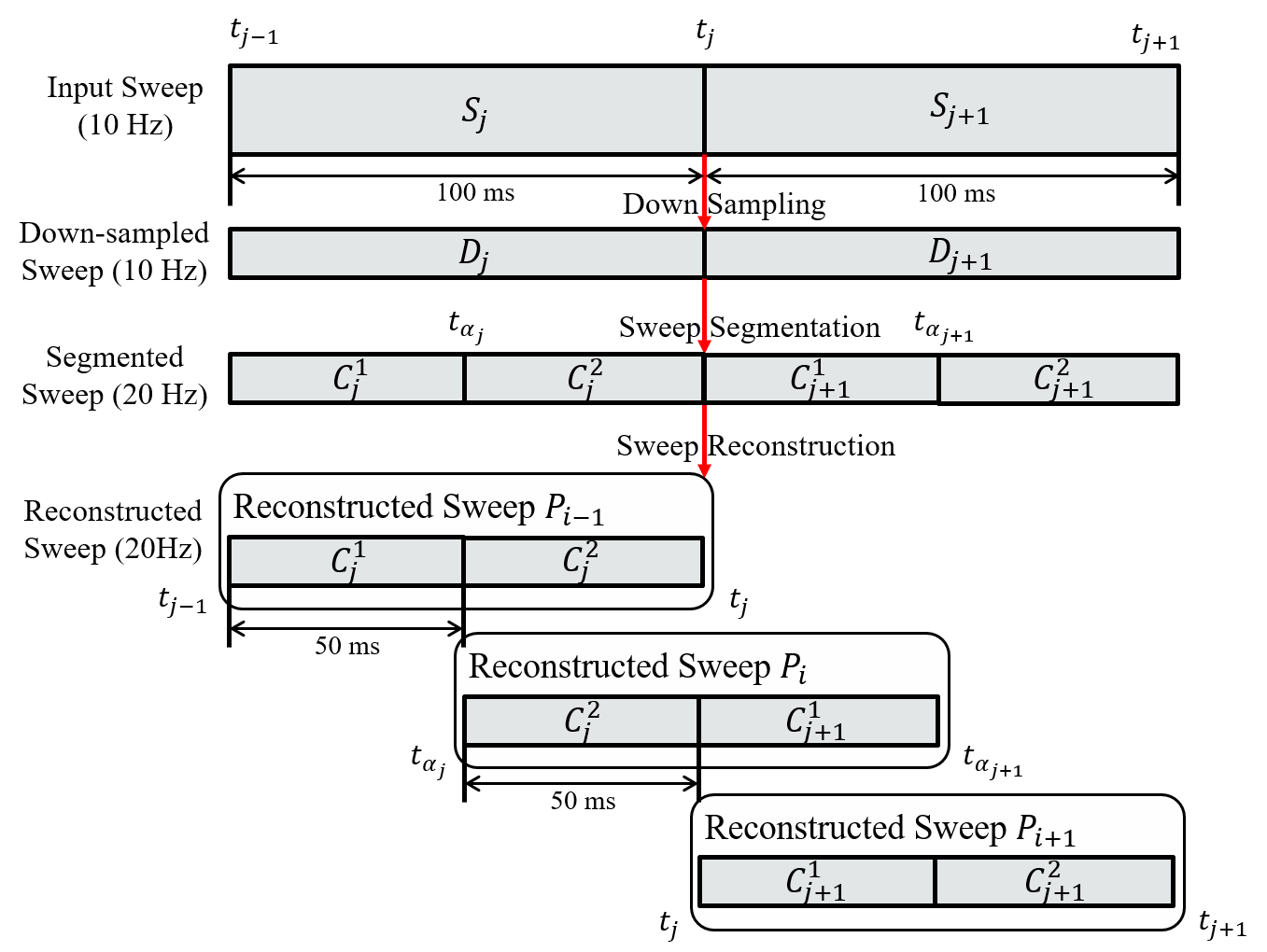}
		\caption{Illustration of sweep reconstruction method, which splits the original sweep packet into continuous point cloud data segments, and then re-packages point cloud data streams in a multiplexing way to obtain sweeps with higher frequency.}
		\label{fig1}
	\end{center}
\end{figure}

In this paper, we found that our previous proposed sweep reconstruction method can reduce the accumulative error of predicted state by reducing the time interval of IMU measurements integration for iEKF based LIO systems by using high-frequent reconstructed sweeps, and in turn achieve more accurate and robust state estimation results. Specifically, the sweep reconstruction method uses the characteristics of continuous scanning of spinning LiDAR to segment and reconstruct raw input sweeps from spinning LiDAR to obtain reconstructed sweeps with higher frequency (as shown in Fig. \ref{fig1}). The increased frequency shortens the time interval between two consecutive sweeps, thus reduces the time interval of IMU measurements integration and increasing the frequency of state update. Therefore, the sweep reconstruction can not only increase the frequency of output pose, but also improve the accuracy of state estimation of iEKF based LIO. In addition, to prevent inaccurate trajectory caused by multiple inconsistent distortion correction to a particular point, we further propose to perform distortion correction for each segment, which ensured the accuracy of estimated trajectory and map.

We integrate the sweep reconstruction method and the corresponding distortion correction method into our iEKF-based LIO system to derived our SR-LIO. Experimental results on four public datasets demonstrate that: 1) our SR-LIO outperforms all existing state-of-the-art methods on accuracy and achieves higher frequent output pose; 2) reducing the time interval of iterated state update via the proposed sweep reconstruction can improve the accuracy of estimated states; 3) the distortion correction for segments can better ensure the accuracy of estimated trajectory.

To summarize, the main contributions of this work are three folds: 1) We embed the previous proposed sweep reconstruction method into our newly designed iEKF based LIO system and achieve the state-of-the-art accuracy; 2) For reconstructed sweeps, we proposed an undistort strategy which performs distortion correction for each segment. 3) We have released the source code of this work for the development of the community\footnote{https://github.com/ZikangYuan/sr\_lio}.

The rest of this paper is structured as follows. In Sec. \ref{Related Work}, we briefly discuss the relevant literature. Sec. \ref{Preliminary} provides preliminaries. Then Secs. \ref{Our System SR-LIO} presents details of our system SR-LIO. Sec. \ref{Experiments} provides experimental evaluation. Finally, we conclude the paper in Sec. \ref{Conclusion}.

\section{Related Work}
\label{Related Work}

\textbf{LiDAR-Only Odometry.} LiDAR-based odometry systems \cite{zhang2014loam, zhang2017low, shan2018lego, behley2018efficient, wang2021f, deschaud2018imls, dellenbach2022ct} rely on geometric information contained in LiDAR point clouds for tracking, and constantly register the new point cloud to the map. LOAM \cite{zhang2014loam, zhang2017low} is the most classical LiDAR odometry, which mainly consists of three steps: 1) Extracting edge and surface features from raw point clouds; 2) Performing sweep-to-sweep pose estimation; 3) Performing sweep-to-map pose optimization and utilizing the optimized pose to register point clouds to the map. However, due to huge number of 3D point clouds to be processed, the output frequency of LOAM is low. On the basis of LOAM, LeGO-LOAM \cite{shan2018lego} proposes to group raw point clouds into several clusters, and removes clusters with weak geometric structure information to reduce computation. However, accurately removing clusters with weak geometry is a nontrivial task, and incorrect removal of useful clusters would degrade the accuracy and robustness of pose estimation. SuMa \cite{behley2018efficient} proposes to represent the map via a surfel-based representation that aggregates information from point clouds. However, GPU acceleration is necessary for SuMa to achieve real-time performance, and the pose estimation accuracy of SuMa is not better than systems based on the framework of LOAM. Fast-LOAM \cite{wang2021f} eliminates the sweep-to-sweep pose estimation module and only retains the sweep-to-map pose estimation module. Meanwhile, Fast-LOAM utilized analytic derivative instead of automatic derivation to speed up the Ceres Solver \cite{Agarwal_Ceres_Solver_2022}. IMLS-LOAM \cite{deschaud2018imls} proposes an IMLS pose solution algorithm to replace conventional ICP. However, large computational cost of IMLS makes IMLS-LOAM impossible to run in real time. CT-ICP \cite{dellenbach2022ct} is the first open-sourced elastic LiDAR odometry framework, where the state at the beginning of current sweep and the state at the end of current sweep are variables to be optimized, and a logical constraint is added to make the beginning state approaches to the end state of last sweep. This sweep expression enables to perform pose estimation and the distortion calibration simultaneously. However, an inappropriate weight of logical constraint would result in serious inconsistency of estimated trajectory.

\textbf{LiDAR-Inertial Odometry.} LiDAR-inertial odometry systems \cite{zhang2014loam, zhang2017low, shan2018lego, li2021towards, ye2019tightly, qin2020lins, shan2020lio, xu2021fast, xu2022fast, chen2023direct} are mainly divided into loosely-coupled framework and tightly-coupled framework. The loose-coupled framework, e.g., LOAM \cite{zhang2014loam, zhang2017low} and LeGO-LOAM \cite{shan2018lego} which have an IMU interface, uses IMU measurements to calibrate the motion distortion of LiDAR point clouds, and provides motion priors for ICP pose estimation. The tightly-coupled framework \cite{li2021towards, ye2019tightly, qin2020lins, shan2020lio, xu2021fast, xu2022fast, chen2023direct} uses IMU measurements to provide motion constraints together with ICP, so as to improve pose estimation accuracy and robustness. According to the type of LiDAR-inertial state estimation, the tightly-coupled system can be further divided into iEKF based framework \cite{qin2020lins, xu2021fast, xu2022fast, chen2023direct}, BA based framework \cite{li2021towards, ye2019tightly} and graph optimization based framework \cite{shan2020lio}. LINs firstly fuses 6-axis IMU and 3D LiDAR in an EKF based framework, where an iEKF is designed to correct the estimated state recursively by generating new feature correspondences in each iteration, and to keep the system computationally tractable. Based on the mathematical derivation, Fast-LIO \cite{xu2021fast} adapts a technique of solving Kalman gain \cite{sorenson1966kalman} to avoid the calculation of the high-order matrix inversion, and in turn greatly reduce the computational burden. Based on Fast-LIO, Fast-LIO2 \cite{xu2022fast} proposes an ikd-tree algorithm \cite{cai2021ikd}. Compared with the original kd-tree, ikd-tree reduces time cost in building a tree, traversing a tree, removing elements and other operations. DLIO \cite{chen2023direct} proposes to retain a 3-order minimum in state prediction and point distortion calibration to obtain more accurate pose estimation result. LIO-SAM \cite{shan2020lio} firstly formulates LIO odometry as a factor graph. Such formulation allows a multitude of relative and absolute measurements, including loop closures, to be incorporated from different sources as factors into the system. \cite{ye2019tightly} firstly fuses 6-axis IMU and 3D LiDAR in a BA based framework. Besides, to obtain more reliable poses estimation, a rotation-constrained refinement algorithm is proposed to further align the pose with the global map. LiLi-OM \cite{li2021towards} selects the key-sweeps from solid-state LiDAR data, and performs BA based multi-key-sweep joint LIO-optimization. However, when the type of LiDAR changes from solid-state to spinning, the time interval between two consecutive key-sweeps becomes longer than 100ms, then the error of IMU constraints in LiLi-OM is larger than that of \cite{ye2019tightly}.

\section{Preliminary}
\label{Preliminary}

\subsection{Coordinate Systems}
\label{Coordinate Systems}

We denote $(\cdot)^w$, $(\cdot)^l$ and $(\cdot)^b$ as a 3D point in the world coordinates, the LiDAR coordinates and the IMU coordinates respectively. The world coordinate is coinciding with $(\cdot)^b$ at the starting position.

We denote the LiDAR coordinates for taking the $i_{th}$ sweep at time $t_i$ as $l_i$ and the corresponding IMU coordinates at $t_i$ as $b_i$, then the transformation matrix (i.e., external parameters) from $l_i$ to $b_i$ is denoted as $\mathbf{T}_{l_i}^{b_i} \in S E(3)$, which consists of a rotation matrix $\mathbf{R}_{l_i}^{b_i} \in S O(3)$ and a translation vector $\mathbf{t}_{l_i}^{b_i} \in \mathbb{R}^3$. The external parameters are usually calibrated once offline and remain constant during online pose estimation. Therefore, we can represent $\mathbf{T}_{l_i}^{b_i}$ using $\mathbf{T}_{l}^{b}$ for simplicity. In the following statement, we omit the index that represents the coordinate system for simplified notation. For instance, the pose from the IMU coordinates to the world coordinate is strictly defined as $\mathbf{T}_{w}^{b_i}$.

In addition to pose, we also estimate the velocity $\mathbf{v}$, the accelerometer bias $\mathbf{b}_{\mathbf{a}}$, the gyroscope bias $\mathbf{b}_{\mathbf{\omega}}$ and the gravitational acceleration $\mathbf{g}_{w}$, which are represented uniformly by a state vector:
\begin{equation}
\label{equation1}
	\boldsymbol{x}=\left[\mathbf{t}^T, \mathbf{q}^T, \mathbf{v}^T, {\mathbf{b}_{\mathbf{a}}}^T, {\mathbf{b}_{\boldsymbol{\omega}}}^T, {\mathbf{g}^{w}}^T \right]^T
\end{equation}
where $\mathbf{q}$ is the quaternion form of the rotation matrix $\mathbf{R}$.

\subsection{IMU Measurement Model}
\label{IMU Measurement Model}

An IMU consists of an accelerometer and a gyroscope. The raw gyroscope and accelerometer measurements from IMU, $\hat{\mathbf{a}}_t$ and $\hat{\boldsymbol{\omega}}_t$, are given by:
\begin{equation}
\label{equation2}
	\begin{gathered}
		\hat{\mathbf{a}}_t=\mathbf{a}_t+\mathbf{b}_{\mathbf{a}_t}+\mathbf{R}_w^t \mathbf{g}^w+\mathbf{n}_{\mathbf{a}} \\
		\hat{\boldsymbol{\omega}}_t=\boldsymbol{\omega}_t+\mathbf{b}_{\boldsymbol{\omega}_t}+\mathbf{n}_{\boldsymbol{\omega}}
	\end{gathered}
\end{equation}
IMU measurements, which are measured in the IMU coordinates, combine the force for countering gravity and the platform dynamics, and are affected by acceleration bias $\mathbf{b}_{\mathbf{a}_t}$, gyroscope bias $\mathbf{b}_{\boldsymbol{\omega}_t}$, and additive noise. As mentioned in VINs-Mono \cite{qin2018vins}, the additive noise in acceleration and gyroscope measurements can be modeled as Gaussian white noise, $\mathbf{n}_{\mathbf{a}} \sim N\left(\mathbf{0}, \boldsymbol{\sigma}_{\mathbf{a}}^2\right)$, $\mathbf{n}_{\boldsymbol{\omega}} \sim N\left(\mathbf{0}, \boldsymbol{\sigma}_{\boldsymbol{\omega}}^2\right)$. Acceleration bias and gyroscope bias are modeled as random walk, whose derivatives are Gaussian, $\dot{\mathbf{b}}_{\mathbf{a}_t}=\mathbf{n}_{\mathbf{b}_{\mathbf{a}}} \sim N\left(\mathbf{0}, \boldsymbol{\sigma}_{\mathbf{b}_{\mathbf{a}}}^2\right)$, $\dot{\mathbf{b}}_{\boldsymbol{\omega}_t}=\mathbf{n}_{\mathbf{b}_{\boldsymbol{\omega}}} \sim N\left(\mathbf{0}, \boldsymbol{\sigma}_{\mathbf{b}_{\boldsymbol{\omega}}}^2\right)$.

\section{Our System SR-LIO}
\label{Our System SR-LIO}

\subsection{Overview}
\label{Overview}

\begin{figure}
	\begin{center}
		\includegraphics[scale=0.47]{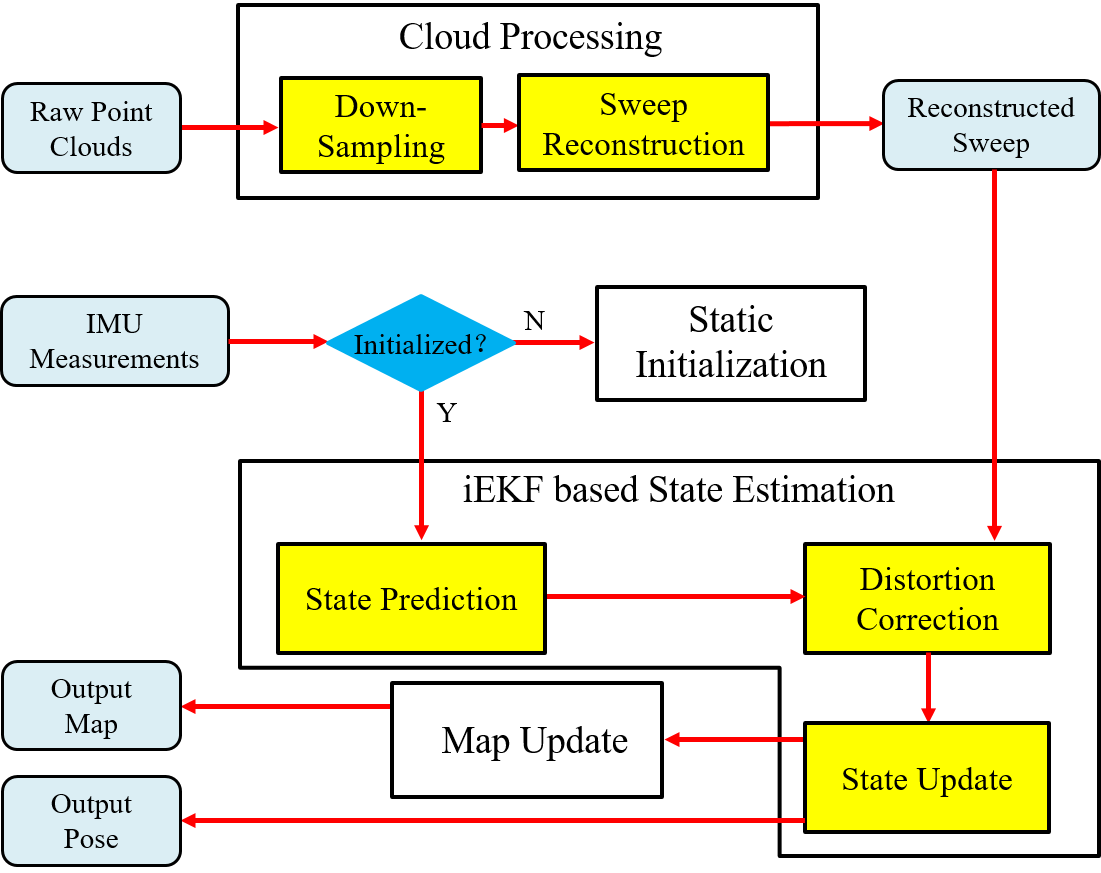}
		\caption{Overview of our SR-LIO which consists of four main modules: a cloud processing module, a static initialization module, an iEKF based State Estimation module and a map update module.}
		\label{fig2}
	\end{center}
\end{figure}
Fig. \ref{fig2} illustrates the framework of our SR-LIO which consists of four main modules: cloud processing, static initialization, iEKF based state estimation and map update. The cloud processing module down-samples the 10 Hz input sweep, then segments and reconstructs the 10 Hz down-sampled sweep to obtain a reconstructed sweep at 20 Hz. The static initialization module utilizes the IMU measurements to estimate some state parameters such as gravitational acceleration, accelerometer bias, gyroscope bias, and initial velocity. The iEKF based state estimation module perform state estimation in real time, where the frequency of state update is equal to the frequency of reconstructed sweep. Finally, we add the point clouds to the map and delete the point clouds that are far away. For map management, we utilized the Hash voxel map, which is the same as CT-ICP \cite{dellenbach2022ct}.

\subsection{Cloud Processing}
\label{Cloud Processing}

\subsubsection{Down-Sampling}
\label{Down-Sampling}

Due to huge number of 3D point clouds to be processed, the computational burden of the whole system is heavy. In order to reduce the computational burden, we down-sample the input point clouds (i.e., $S_j$ and $S_{j+1}$ in Fig. \ref{fig1}). Firstly, we perform the quantitative down-sampling strategy, which keeps only one out of every four points. Then, we put the quantitative down-sampled points into a volume with $0.5\times0.5\times0.5$ (unit: m) voxel size, and make each voxel contain only one point cloud, which is the same as CT-ICP \cite{dellenbach2022ct}.

\subsubsection{Sweep Reconstruction}
\label{Sweep Reconstruction}

Sweep reconstruction aims to derive a 20\,Hz reconstructed sweep $P$ from the 10\,Hz original input point cloud $S$. Fig. \ref{fig1} illustrates the core idea of sweep reconstruction, which is proposed in our previous work \cite{yuan2023sdv}. Given the last sweep $S_j$ which begins at $t_{j-1}$ and ends at $t_j$, and the current input sweep $S_{j+1}$ which begins at $t_j$ and ends at $t_{j+1}$, we assume the lengths of time intervals $\left[t_{j-1}, t_j\right]$ and $\left[t_j, t_{j+1}\right]$ are both 100\,ms. Based on the characteristics of continuous acquisition over a period of time of LiDAR, we can split the original sweep packet into continuous point cloud data streams, and then re-package point cloud data streams in a multiplexing way to obtain sweeps with higher frequency. Specifically, we first calculate two equal points of the time interval $\left[t_{j-1}, t_j\right]$ and $\left[t_j, t_{j+1}\right]$ (i.e., $t_{\alpha_j}$ and $t_{\alpha_{j+1}}$), and put all time stamps in a set:
\begin{equation}
\label{equation3}
	T=\left\{t_{j-1}, t_{\alpha_j}, t_j, t_{\alpha_{j+1}}, t_{j+1}\right\}
\end{equation}
Then we take each element of $T$ (i.e., $T\left[k\right]$) as the beginning time stamp and take $T\left[k+2\right]$ as the end time stamp. We re-packet the point cloud data streams during $\left[T\left[k\right], T\left[k+2\right]\right]$ to obtain the reconstructed sweep. For instance, we packet the point cloud data streams during $\left[t_{\alpha_j}, t_{\alpha_{j+1}}\right]$ to obtain the reconstructed sweep $P_i$. By this way, the original sweeps $S_{j+1}$ can be re-packing to obtain two reconstructed sweeps (e.g., $P_i$ and $P_{i+1}$). Although the duration of $P_i$ is still 100\,ms, the time interval between two constructed sweeps (e.g., $P_i$ and $P_{i+1}$) decreases from 100\,ms to 50\,ms. Therefore, our proposed sweep reconstruction can increase the frequency of sweep from 10\,Hz to 20\,Hz.

\subsection{Static Initialization}
\label{Static Initialization}

We adopt static initialization \cite{geneva2020openvins} in our system to estimate some necessary variables including initial velocity, gravitational acceleration, accelerometer bias and gyroscope bias. Please refer to \cite{geneva2020openvins} for more details.

\subsection{iEKF based State Estimation}
\label{iEKF based State Estimation}

The same as Fast-LIO \cite{xu2021fast}, we utilize the iterated extended Kalman filter (iEKF) to perform state estimation. We set the error state
\begin{equation}
\label{equation4}
	\delta \boldsymbol{x}=\left[\delta \mathbf{t}, \delta \boldsymbol{\theta}, \delta \mathbf{v}, \delta \mathbf{b}_{\mathbf{a}}, \delta \mathbf{b}_{\boldsymbol{\omega}}, \delta \mathbf{g}\right]^T
\end{equation}
as the state variable of the filter to derive the prediction and update formula. It is necessary to note that $\delta \boldsymbol{\theta} \in so(3)$, which is the Lie algebra of rotation. $\delta \mathbf{t}$, $\delta \mathbf{v}$, $\delta \mathbf{b}_{\mathbf{a}}$, $\delta \mathbf{b}_{\boldsymbol{\omega}} \in \mathbb{R}^3$, $\delta \mathbf{g} \in S^2$ due to the fixed length of gravitational acceleration. The estimated error state would be added to the optimal state (Eq. \ref{equation1}) during each iteration of state update.

\subsubsection{State Prediction}
\label{State Prediction}

The state prediction is performed once receiving an IMU input (i.e., $\hat{\boldsymbol{\omega}}_{n+1}$ and $\hat{\mathbf{a}}_{n+1}$), while the optimal state $\boldsymbol{x}_{n+1}^w$ (i.e., $\mathbf{t}_{n+1}^w$, $\mathbf{R}_{n+1}^w$, $\mathbf{v}_{n+1}^w$, $\mathbf{b}_{\mathbf{a}_{n+1}}$, $\mathbf{b}_{\boldsymbol{\omega}_{n+1}}$, $\mathbf{g}_{n+1}^w$) is calculated by:
\begin{equation}
\label{equation5}
	\begin{gathered}
		\mathbf{R}_{n+1}^w=\mathbf{R}_n^w Exp\left(\left(\frac{\hat{\boldsymbol{\omega}}_n+\hat{\boldsymbol{\omega}}_{n+1}}{2}-\mathbf{b}_{\boldsymbol{\omega}_n}\right) \Delta t\right) \\
		\mathbf{v}_{n+1}^w=\mathbf{v}_n^w+\left(\frac{\hat{\mathbf{a}}_n+\hat{\mathbf{a}}_{n+1}}{2}-\mathbf{b}_{\mathbf{a}_n}-\mathbf{R}_w^n \mathbf{g}_n^w\right) \Delta t \\
		\mathbf{t}_{n+1}^w=\mathbf{t}_n^w+\mathbf{v}_n^w \Delta t+\frac{1}{2}\left(\frac{\hat{\mathbf{a}}_n+\hat{\mathbf{a}}_{n+1}}{2}-\mathbf{b}_{\mathbf{a}_n}-\mathbf{R}_w^n \mathbf{g}_n^w\right) \Delta t^2 \\
		\mathbf{b}_{\mathbf{a}_{n+1}}=\mathbf{b}_{\mathbf{a}_n}, \mathbf{b}_{\boldsymbol{\omega}_{n+1}}=\mathbf{b}_{\boldsymbol{\omega}_n}, \mathbf{g}_{n+1}^w=\mathbf{g}_n^w
	\end{gathered}
\end{equation}
The error state $\delta \boldsymbol{x}_{n+1}$ and covariance $P_{n+1}$ is propagated as:
\begin{equation}
\label{equation6}
	\begin{gathered}
		\delta \boldsymbol{x}_{n+1}=\mathbf{F}_{\boldsymbol{x}} \delta \boldsymbol{x}_n \\
		\mathbf{P}_{n+1}=\mathbf{F}_{\boldsymbol{x}} \mathbf{P}_n {\mathbf{F}_{\boldsymbol{x}}}^T+\mathbf{F}_{\boldsymbol{w}} \mathbf{Q} {\mathbf{F}_{\boldsymbol{w}}}^T
	\end{gathered}
\end{equation}
where $\mathbf{Q}$ is the diagonal covariance matrix of noise ($\boldsymbol{\sigma}_{\mathbf{a}}^2$, $\boldsymbol{\sigma}_{\boldsymbol{\omega}}^2$, $\boldsymbol{\sigma}_{\mathbf{b}_{\mathbf{a}}}^2$, $\boldsymbol{\sigma}_{\mathbf{b}_{\boldsymbol{\omega}}}^2$). $\Delta t$ is the time interval between two consecutive IMU measurements. $\mathbf{F}_{\boldsymbol{x}}$ is expressed as:
\begin{equation}
\label{equation7}
	\mathbf{F}_{\boldsymbol{x}}=\left[\begin{array}{cccccc}
		\mathbf{I} & \mathbf{0} & \mathbf{I} \Delta t & \mathbf{0} & \mathbf{0} & \mathbf{0} \\
		\mathbf{0} & f_{11} & \mathbf{0} & \mathbf{0} & -\mathbf{I} \Delta t & \mathbf{0} \\
		\mathbf{0} & f_{21} & \mathbf{I} & -\mathbf{R}_w^n \Delta t & \mathbf{0} & f_{25} \\
		\mathbf{0} & \mathbf{0} & \mathbf{0} & \mathbf{I} & \mathbf{0} & \mathbf{0} \\
		\mathbf{0} & \mathbf{0} & \mathbf{0} & \mathbf{0} & \mathbf{I} & \mathbf{0} \\
		\mathbf{0} & \mathbf{0} & \mathbf{0} & \mathbf{0} & \mathbf{0} & f_{55}
	\end{array}\right]
\end{equation}
\begin{equation}
\label{equation8}
	f_{11}=\mathbf{I}-\left(\frac{\hat{\boldsymbol{\omega}}_n+\hat{\boldsymbol{\omega}}_{n+1}}{2}-\mathbf{b}_{\boldsymbol{\omega}_n}\right) \Delta t
\end{equation}
\begin{equation}
\label{equation9}
	f_{21}=-\mathbf{R}_w^n\left(\frac{\hat{\mathbf{a}}_n+\hat{\mathbf{a}}_{n+1}}{2}-\mathbf{b}_{\mathbf{a}_n}\right)^{\wedge} \Delta t
\end{equation}
\begin{equation}
\label{equation10}
	f_{25}=\mathbf{g}_n^{w^{\wedge}} \mathbf{B}\left(\mathbf{g}_n^w\right) \Delta t
\end{equation}
\begin{equation}
\label{equation11}
	f_{55}=-\frac{1}{\left\|\mathbf{g}_n^w\right\|^2} \mathbf{B}\left(\mathbf{g}_n^w\right)^T \mathbf{g}_n^{w \wedge} \mathbf{g}_n^{w \wedge} \mathbf{B}\left(\mathbf{g}_n^w\right)
\end{equation}
\begin{equation}
\label{equation12}
	\mathbf{B}(\mathbf{g})=\left[\begin{array}{cc}
		1-\frac{\overline{\mathbf{g}}_x^2}{1+\overline{\mathbf{g}}_z} & -\frac{\overline{\mathbf{g}}_x \overline{\mathbf{g}}_y}{1+\overline{\mathbf{g}}_z} \\
		-\frac{\overline{\mathbf{g}}_x \overline{\mathbf{g}}_y}{1+\overline{\mathbf{g}}_z} & 1-\frac{\overline{\mathbf{g}}_y{ }^2}{1+\overline{\mathbf{g}}_z} \\
		-\overline{\mathbf{g}}_x & -\overline{\mathbf{g}}_y
	\end{array}\right]
\end{equation}
where $(\overline{\cdot})$ indicates normalization for a specific element and $(\cdot)^{\wedge}$ indicates the skew symmetric matrix corresponding to a vector. $\mathbf{F}_{\boldsymbol{w}}$ is expressed as:
\begin{equation}
\label{equation13}
	\mathbf{F}_{\boldsymbol{w}}=\left[\begin{array}{cccc}
		\mathbf{0} & \mathbf{0} & \mathbf{0} & \mathbf{0} \\
		\mathbf{0} & -\mathbf{I} \Delta t & \mathbf{0} & \mathbf{0} \\
		-\mathbf{R}_w^n \Delta t & \mathbf{0} & \mathbf{0} & \mathbf{0} \\
		\mathbf{0} & \mathbf{0} & -\mathbf{I} \Delta t & \mathbf{0} \\
		\mathbf{0} & \mathbf{0} & \mathbf{0} & -\mathbf{I} \Delta t \\
		\mathbf{0} & \mathbf{0} & \mathbf{0} & \mathbf{0}
	\end{array}\right]
\end{equation}
It is not difficult to see from Eq. \ref{equation6} that: the uncertainty (expressed by covariance $\mathbf{P}$) of the predicted state increases with the increase of IMU measurements being integrated. Therefore, the longer time interval between two state update is, the larger accumulative error exists in the predicted state.

\subsubsection{Distortion Correction}
\label{Distortion Correction}

\begin{figure}
	\begin{center}
		\includegraphics[scale=0.365]{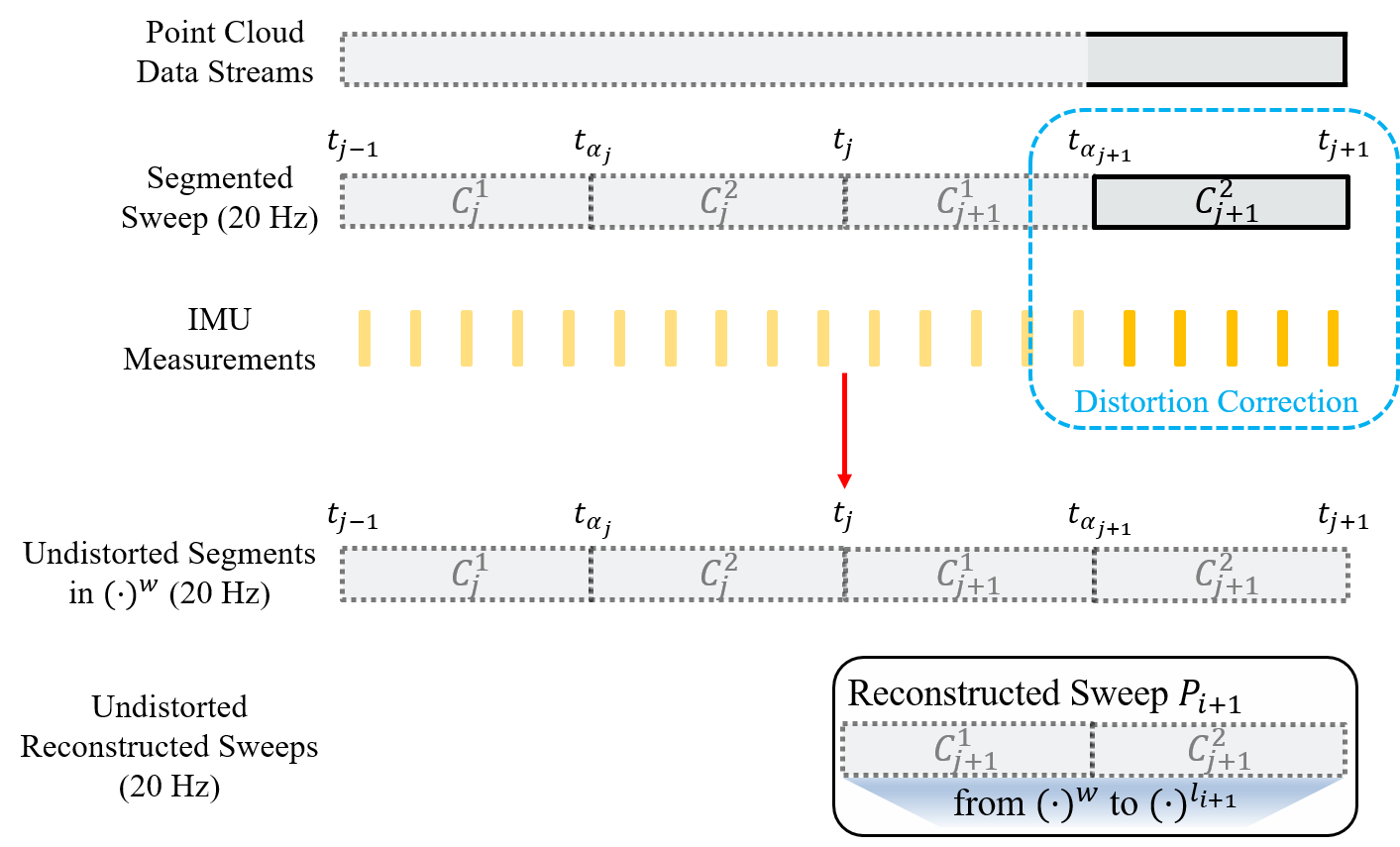}
		\caption{Illustration of distortion correction. We perform distortion correction for each segment, but not an entire reconstructed sweep. This strategy makes the specific point to be undistorted only once, to ensure the consistency of state estimation.}
		\label{fig3}
	\end{center}
\end{figure}

There are two options to perform distortion correction: 1) Performing distortion correction for each reconstructed sweep; 2) Performing distortion correction for each segment. The option 1) causes the point cloud in a specific period to be undistorted multi times. Since the pose used for each distortion correction are different, for a specific point, the coordinate in $(\cdot)^w$ after two corrections will not be the same. This problem would result in the inaccuracy of estimated trajectory. To solve this problem, we propose to perform distortion correction for each segment, but not an entire sweep (as shown in Fig. \ref{fig3}). Specifically, for each segment (e.g., $C_{j+1}^2$), we transform the point (e.g., $\mathbf{p} \in C_{j+1}^2$) to $(\cdot)^w$ according to IMU-integrated pose or the uniform motion model. After obtaining the reconstructed sweep $P_{i+1}$, we transform all points belong to it from $(\cdot)^w$ to $(\cdot)^{l_{i+1}}$ to finish the distortion correction.

\subsubsection{State Update}
\label{State Update}

When every new reconstructed sweep $P_{i+1}$ completes, we iteratively perform the following steps for state prediction.

\textbf{Step1. Point-to-plane residuals computation.} During each iteration, we firstly build the point-to-plane residuals. Specifically, for a undistorted point-to-plane residuals. Specifically, for a undistorted point $\mathbf{p}_{k} \in P_{i+1}$ $(1 \leq \mathrm{k} \leq \mathrm{m})$, we first project $\mathbf{p}_k$ to the world coordinate to obtain ${\mathbf{p}_k}^w$, and then find 20 nearest points around ${\mathbf{p}_k}^w$ from the volume. To search for the nearest neighbor of ${\mathbf{p}_k}^w$, we only search in the voxel $V$ to which ${\mathbf{p}_k}^w$ belongs, and the 8 voxels adjacent to $V$. The 20 nearest points are used to fit a plane with a normal $\mathbf{n}$ and a distance $d$. Accordingly, we can build the point-to-plane residual $r^{\mathbf{p}_k}$ for $\mathbf{p}_k$ as the observation constraint:
\begin{equation}
\label{equation14}
	\begin{aligned}
		& r^{\mathbf{p}_k}=\omega_{\mathbf{p}}\left(\mathbf{n}^T \mathbf{p}_k{ }^w+d\right) \\
		& {\mathbf{p}_k}^w=\mathbf{R}_{b_{i+1}}^w \mathbf{p}_k+\mathbf{t}_{b_{i+1}}^w
	\end{aligned}
\end{equation}
where $\omega_{\mathbf{p}}$ is a weight parameter utilized in \cite{dellenbach2022ct}, $\mathbf{R}_{b_{i+1}}^w$ is the rotation from $(\cdot)^{b_{i+1}}$ to $(\cdot)^w$ at $t_{i+1}$. We can express the observation matrix $\mathbf{h}$ as:
\begin{equation}
\label{equation15}
	\mathbf{h}=\left[{r^{\mathbf{p}_1}}^T, {r^{\mathbf{p}_2}}^T, \cdots, {r^{\mathbf{p}_m}}^T\right]^T
\end{equation}
The corresponding Jacobin matrix of observation constraint $\mathbf{H}$ is calculated as:
\begin{equation}
\label{equation16}
	\begin{gathered}
		\mathbf{H}=\left[{\mathbf{H}_1}^T, {\mathbf{H}_2}^T, \cdots, {\mathbf{H}_m}^T\right]^T \\
		\mathbf{H}^T=\left[\begin{array}{llllll}
			\omega_{\mathbf{p}} \mathbf{n}^T & -\omega_{\mathbf{p}} \mathbf{n}^T \mathbf{R}_{b_{i+1}}^w {\mathbf{p}_k}^{\wedge} & \mathbf{0} & \mathbf{0} & \mathbf{0} & \mathbf{0}
		\end{array}\right]^T
	\end{gathered}
\end{equation}
The corresponding covariance matrix of observation constraint $\mathbf{V}$ is the diagonal matrix of ($\mathbf{V}_{1}$, $\mathbf{V}_{2}$, $\cdots$, $\mathbf{V}_{m}$), while $\mathbf{V}_{k}=0.001$ in our system.

\textbf{Step2. Incremental computation.} We define the optimal state calculated from state prediction is $\left.\boldsymbol{x}_{b_{i+1}}^w\right|_0$, and the optimal state before current iteration is $\left.\boldsymbol{x}_{b_{i+1}}^w\right|_n$. According to the formula of state update, the incremental $\delta \boldsymbol{x}$ is calculated as:
\begin{equation}
\label{equation17}
	\begin{gathered}
		\mathbf{K}=\left(\mathbf{H}^T \mathbf{R}^{-1} \mathbf{H}+\left(\mathbf{J}_n^0 \mathbf{P} {\mathbf{J}_n^{0}}^T\right)^{-1}\right)^{-1} \mathbf{H}^T \mathbf{V}^{-1} \\
		\delta \boldsymbol{x}=-\mathbf{K h}-(\mathbf{I}-\mathbf{K H}) \mathbf{J}_n^0\left(\left.\left.\boldsymbol{x}_{b_{i+1}}^w\right|_n \boxminus \boldsymbol{x}_{b_{i+1}}^w\right|_0\right)
	\end{gathered}
\end{equation}
where $\mathbf{J}_n^0$ is the partial differentiation of $\left.\left(\boldsymbol{x}_{b_{i+1}}^w|_n \boxplus \delta \boldsymbol{x}\right) \boxminus \boldsymbol{x}_{b_{i+1}}^w\right|_0$ with respect to $\delta \boldsymbol{x}$ evaluated at zero:
\begin{equation}
\label{equation18}
	\mathbf{J}_n^0=\left[\begin{array}{cccc}
		\mathbf{I}_{3 \times 3} & \mathbf{0}_{3 \times 3} & \mathbf{0}_{3 \times 9} & \mathbf{0}_{3 \times 2} \\
		\mathbf{0}_{3 \times 3} & \mathbf{I}-\frac{1}{2} \delta \boldsymbol{\theta}_n^0 & \mathbf{0}_{3 \times 9} & \mathbf{0}_{3 \times 2} \\
		\mathbf{0}_{9 \times 3} & \mathbf{0}_{9 \times 3} & \mathbf{I}_{9 \times 9} & \mathbf{0}_{9 \times 2} \\
		\mathbf{0}_{2 \times 3} & \mathbf{0}_{2 \times 3} & \mathbf{0}_{2 \times 9} & \mathbf{J}_{\mathbf{g}_n}^0
	\end{array}\right]
\end{equation}
\begin{equation}
\label{equation19}
	\mathbf{J}_{\mathbf{g}_n}^0=\mathbf{I}+\frac{1}{2} \mathbf{B}\left(\left.\mathbf{g}_{i+1}^w\right|_0\right)^T {\delta \boldsymbol{\theta}_{\mathbf{g}_{i+1}^w}}^{\wedge} \mathbf{B}\left(\left.\mathbf{g}_{i+1}^w\right|_0\right)
\end{equation}
$\delta \boldsymbol{\theta}_n^0=\mathrm{Log} \left({\left.\mathbf{R}_{b_{i+1}}^w\right|_0}^T \left.\mathbf{R}_{b_{i+1}}^w\right|_n\right)$, $\delta \boldsymbol{\theta}_{\mathbf{g}_{i+1}^w}$ is the Lie algebra of rotation from $\left.\mathbf{g}_{i+1}^w\right|_n$ to $\left.\mathbf{g}_{i+1}^w\right|_0$.

The definition of $\boxminus$: For the variable of type $\mathbf{a}, \mathbf{b} \in \mathbb{R}^3$, $\mathbf{a} \boxminus \mathbf{b}=\mathbf{a}-\mathbf{b}$. For the variable of type $\mathbf{R}_1, \mathbf{R}_2 \in SO(3)$, $\mathbf{R}_1 \boxminus \mathbf{R}_2 = \mathrm{Log}({\mathbf{R}_2}^T\mathbf{R}_1)$. For the variable of type $\mathbf{g}_1, \mathbf{g}_2 \in \mathbb{R}^3$, $\delta \mathbf{g} \in S^2$, $\delta \mathbf{g} = \mathbf{g}_1 \boxminus \mathbf{g}_2 = \mathbf{B}{(\mathbf{g}_2)}^T \delta \boldsymbol{\theta}_{\mathbf{g}}$, where $\boldsymbol{\theta}_{\mathbf{g}}$ is the Lie algebra of rotation from $\mathbf{g}_1$ to $\mathbf{g}_2$.

The definition of $\boxplus$: For the variable of type $\mathbf{a}, \mathbf{b} \in \mathbb{R}^3$, $\mathbf{a} \boxplus \mathbf{b} = \mathbf{a} + \mathbf{b}$. For the variable of type $\mathbf{R} \in SO(3)$ and $\boldsymbol{\theta} \in so(3)$, $\mathbf{R} \boxplus \boldsymbol{\theta} = \mathbf{R} \mathrm{Exp}(\boldsymbol{\theta})$. For the variable of type $\mathbf{g}_1, \mathbf{g}_2 \in \mathbb{R}^3$, $\delta \mathbf{g} \in S^2$, $\mathbf{g}_2 = \mathbf{g}_1 \boxplus \delta \mathbf{g} = \mathrm{Exp}(\mathbf{B}(\mathbf{g}_1) \delta \mathbf{g}) \mathbf{g}_1$.

After the incremental $\delta \boldsymbol{x}$ is calculated, we update the optimal state by:
\begin{equation}
\label{equation20}
	\left.\boldsymbol{x}_{b_{i+1}}^w\right|_{n+1}=\left.\boldsymbol{x}_{b_{i+1}}^w\right|_n \boxplus \delta \boldsymbol{x}
\end{equation}
$\mathbf{Step1}$ and $\mathbf{Step2}$ are performed alternately until one of the following convergence conditions is met: 1) The maximum number of iterations (e.g., 6) was reached. 2) The magnitude of incremental is smaller than a threshold (e.g., 0.1\,degree for rotation and 0.01\,m for translation). After convergence, the covariance is updated as:
\begin{equation}
\label{equation21}
	\mathbf{P}=\mathbf{J}_{n+1}^n(\mathbf{I}-\mathbf{K} \mathbf{H}) \mathbf{P} {\mathbf{J}_{n+1}^n}^T
\end{equation}

\subsection{Map Update}
\label{Map Update}

Following CT-ICP \cite{dellenbach2022ct}, the cloud map is stored in a volume. The size of each voxel is $1.0\times1.0\times1.0$ (unit: m) and each voxel contains a maximum of 20 points. When the state of current down-sampled sweep $P_{i+1}$ has been estimated, we transform $P_{i+1}$ to the world coordinate system $(\cdot)^w$, and add the transformed points into the volume map. If a voxel already has 20 points, the new points cannot be added to it. Meanwhile, we delete the points that are far away from current position.

\section{Experiments}
\label{Experiments}

We evaluated our SR-LIO on four public datasets $nclt$ \cite{carlevaris2016university}, $utbm$ \cite{yan2020eu}, ulhk \cite{wen2020urbanloco} and kaist \cite{jeong2019complex}. $nclt$ is a large-scale, long-term autonomous unmanned ground vehicle dataset collected in the University of Michigans North Campus. The $nclt$ dataset contains a full data stream from a Velodyne HDL-32E LiDAR and 50\,Hz data from Microstrain MS25 IMU. Different from the other three datasets (i.e., $utbm$, $ulhk$ and $kaist$), the LiDAR of nclt takes 130$\sim$140\,ms to complete a 360\,deg sweep (i.e., the frequency of a sweep is about 7.5\,Hz). In addition, at the end of some sequences of $nclt$, the Segway vehicle platform enters a long indoor corridor through a door from the outdoor scene, yielding significant scene changes. This large differences in scenes produce great difficulties for ICL point cloud registration, and hence almost all systems would break down here. Therefore, we omit the test for these cases which usually locate at the end of the sequences. In addition, 50\,Hz IMU measurements cannot meet the requirements of some systems (e.g., LIO-SAM \cite{shan2020lio}). Therefore, we increase the frequency of the IMU to 100\,Hz by interpolation.

\begin{table}[]
\begin{center}
	\caption{Datasets for Evaluation}
	\label{table1}
	\begin{tabular}{c|cc|cc}
		\hline
		\multirow{2}{*}{} & \multicolumn{2}{c|}{Velodyne LiDAR} & \multicolumn{2}{c}{IMU} \\ \cline{2-5} 
		& Type             & Rate             & Type       & Rate       \\ \hline
		$nclt$              & 32               & 7.5              & 9-axis     & 100\,Hz     \\
		$utbm$              & 32               & 10               & 6-axis     & 100\,Hz     \\
		$ulhk$              & 32               & 10               & 9-axis     & 100\,Hz     \\
		$kaist$             & 16               & 10               & 9-axis     & 200\,Hz     \\ \hline
	\end{tabular}
\end{center}
\end{table}

\begin{table}[]
\begin{center}
	\caption{Datasets of All Sequences for Evaluation}
	\label{table2}
	\begin{tabular}{cccc}
		\hline
		& Name         & \begin{tabular}[c]{@{}c@{}}Duration\\ (min:sec)\end{tabular} & \begin{tabular}[c]{@{}c@{}}Distance\\ (km)\end{tabular} \\ \hline
		$nclt\_1$  & 2012-01-08   & 92:16                                                        & 6.4                                                     \\
		$nclt\_2$  & 2012-02-02   & 98:37                                                        & 6.5                                                     \\
		$nclt\_3$  & 2012-02-04   & 77:39                                                        & 5.5                                                     \\
		$nclt\_4$  & 2012-02-05   & 93:40                                                        & 6.5                                                     \\
		$nclt\_5$  & 2012-05-11   & 83:36                                                        & 6.0                                                     \\
		$nclt\_6$  & 2012-05-26   & 97:23                                                        & 6.3                                                     \\
		$nclt\_7$  & 2012-06-15   & 55:10                                                        & 4.1                                                     \\
		$nclt\_8$  & 2012-08-04   & 79:27                                                        & 5.5                                                     \\
		$nclt\_9$  & 2012-08-20   & 88:44                                                        & 6.0                                                     \\
		$nclt\_10$ & 2012-09-28   & 76:40                                                        & 5.6                                                     \\
		$nclt\_11$ & 2012-12-01   & 75:50                                                        & 5.0                                                     \\
		$utbm\_1$  & 2018-07-19   & 15:26                                                        & 4.98                                                    \\
		$utbm\_2$  & 2019-01-31   & 16:00                                                        & 6.40                                                    \\
		$utbm\_3$  & 2019-04-18   & 11:59                                                        & 5.11                                                    \\
		$utbm\_4$  & 2018-07-20   & 16:45                                                        & 4.99                                                    \\
		$utbm\_5$  & 2018-07-13   & 16:59                                                        & 5.03                                                    \\
		$ulhk\_1$  & 2019-01-17   & 5:18                                                         & 0.60                                                    \\
		$ulhk\_2$  & 2019-03-16-1 & 2:30                                                         & 0.55                                                    \\
		$kaist\_1$ & urban\_07    & 9:16                                                         & 2.55                                                    \\
		$kaist\_2$ & urban\_08    & 5.07                                                         & 1.56                                                    \\
		$kaist\_3$ & urban\_13    & 24.14                                                        & 2.36                                                    \\ \hline
	\end{tabular}
\end{center}
\end{table}

The $utbm$ dataset contains two 10\,Hz Velodyne HDL-32E and 100\,Hz Xsens MTi-28A53G25 IMU. For point clouds, we only utilize the data from the left LiDAR. The $ulhk$ dataset contains 10\,Hz LiDAR sweep from Velodyne HDL-32E LiDAR and 100\,Hz IMU data from the 9-axis Xsens MTi-10 IMU. $kaist$ contains two 10\,Hz Velodyne VLP-16, 200\,Hz Ssens MTi-300 IMU. Two 3D LiDARs are tilted by approximately $45^{\circ}$. For point clouds, we utilize the data from both two 3D LiDARs. All the sequences of $utbm$, $ulhk$ and $kaist$ are collected in structured urban areas by a human-driving vehicle. The datasets’ information, including the sensors’ type and data rate, are illustrated in Table \ref{table1}. As all four datasets utilize the vehicle platform, we employ static initialization in our system. Details of all the 21 sequences used in this section, including name, duration, and distance, are listed in Table \ref{table2}. For all four datasets, we utilize the universal evaluation metrics – absolute translational error (ATE) as the evaluation metrics. A consumer-level computer equipped with an Intel Core i7-12700 and 32 GB RAM is used for all experiments.

\subsection{Comparison of the State-of-the-Arts}
\label{Comparison of the State-of-the-Arts}

\begin{table}[]
\begin{center}
	\caption{RMSE of ATE Comparison of State-of-the-art (Unit: m)}
	\label{table3}
	\begin{tabular}{c|cccc|c}
		\hline
		& \begin{tabular}[c]{@{}c@{}}LiLi-\\ OM\end{tabular} & \begin{tabular}[c]{@{}c@{}}LIO-\\ SAM\end{tabular} & \begin{tabular}[c]{@{}c@{}}Fast-\\ LIO2\end{tabular} & DLIO           & Ours           \\ \hline
		$nclt\_1$  & 50.71                                              & 1.85                                               & 3.57                                                 & 3.27           & \textbf{1.34}  \\
		$nclt\_2$  & 91.86                                              & 7.18                                               & 2.00                                                 & \textbf{1.80}  & \textbf{1.80}  \\
		$nclt\_3$  & 92.93                                              & \textbf{2.16}                                      & 2.77                                                 & 5.35           & 2.37           \\
		$nclt\_4$  & 215.91                                             & 2.70                                               & 3.60                                                 & 18.10          & \textbf{1.91}  \\
		$nclt\_5$  & 185.24                                             & $\times$                                                  & 2.46                                                 & 3.14           & \textbf{1.62}  \\
		$nclt\_6$  & 141.83                                             & $\times$                                                  & 2.60                                                 & 12.44          & \textbf{2.10}  \\
		$nclt\_7$  & 50.42                                              & 2.97                                               & 2.37                                                 & 2.98           & \textbf{2.13}  \\
		$nclt\_8$  & 137.05                                             & \textbf{2.26}                                      & 2.59                                                 & 7.84           & 2.70           \\
		$nclt\_9$  & 224.68                                             & 10.68                                              & 4.01                                                 & 2.46           & \textbf{2.11}  \\
		$nclt\_10$ & $\times$                                                  & $\times$                                                  & 2.65                                                 & 7.72           & \textbf{1.67}  \\
		$nclt\_11$ & $\times$                                                  & $\times$                                                  & 4.37                                                 & 3.89           & \textbf{1.61}  \\
		$utbm\_1$  & 67.16                                              & -                                                  & 15.13                                                & 14.25          & \textbf{7.70}  \\
		$utbm\_2$  & 38.17                                              & -                                                  & 21.21                                                & \textbf{13.85} & 16.28          \\
		$utbm\_3$  & 10.70                                              & -                                                  & 10.81                                                & 55.28          & \textbf{8.42}  \\
		$utbm\_4$  & 70.98                                              & -                                                  & 15.20                                                & 18.05          & \textbf{11.12} \\
		$utbm\_5$  & 62.57                                              & -                                                  & 13.24                                                & 14.95          & \textbf{9.14}  \\
		$ulhk\_1$  & $\times$                                                  & 1.68                                               & 1.20                                                 & 2.44           & \textbf{0.93}  \\
		$ulhk\_2$  & \textbf{3.11}                                      & 3.13                                               & 3.24                                                 & $\times$              & 3.21           \\
		$kaist\_1$ & $\times$                                                  & 16.96                                              & 0.88                                                 & \textbf{1.04}  & 1.10           \\
		$kaist\_2$ & $\times$                                                  & $\times$                                                  & 16.27                                                & 1.91           & \textbf{0.92}  \\
		$kaist\_3$ & $\times$                                                  & $\times$                                                  & $\times$                                                    & $\times$              & \textbf{1.36}  \\ \hline
	\end{tabular}
\end{center}
\end{table}

We compare our SR-LIO with four state-of-the-art LIO systems, i.e., LiLi-OM \cite{li2021towards}, LIO-SAM \cite{shan2020lio}, Fast-LIO2 \cite{xu2022fast} and DLIO \cite{chen2023direct}. LiLi-OM selects key-sweeps from raw input sweeps, and joints LiDAR point-to-plane residuals and IMU pre-integration residuals into a BA optimization-based framework. LIO-SAM joints the IMU factor and the LiDAR point-to-plane factor into a graph optimization based framework. Both Fast-LIO2 and DLIO are iEKF framework based LIO systems, where Fast-LIO2 utilizes the ikdtree \cite{cai2021ikd} and DLIO utilizes the nanoflan \cite{blanco2014nanoflann} to manage the map. For a fair comparison, we obtain the results of the above systems based on the source code provided by the authors.

Results in Table \ref{table3} demonstrate that our SR-LIO outperforms state-of-the-arts for almost all sequences in terms of smaller ATE. Although our accuracy is not the best on $nclt\_3$, $ulhk\_2$ and $kaist\_1$, we are very close to the best accuracy. “-” means the corresponding value is not available. LIO-SAM needs 9-axis IMU data as input, while the $utbm$ dataset only provides 6-axis IMU data. Therefore, we cannot provide the results of LIO-SAM on the $utbm$ dataset. “$\times$” means the system fails to run entirety on the corresponding sequence. Except for our system and Fast-LIO2, other systems break down on several sequences, which also demonstrate the robustness of our system.

\begin{table}[]
\begin{center}
	\caption{Output Pose Frequency Comparison}
	\label{table4}
	\begin{tabular}{c|cccc|c}
		\hline
		& \begin{tabular}[c]{@{}c@{}}LiLi-\\ OM\end{tabular} & \begin{tabular}[c]{@{}c@{}}LIO-\\ SAM\end{tabular} & LINs & \begin{tabular}[c]{@{}c@{}}Fast-\\ LIO2\end{tabular} & Ours        \\ \hline
		nclt  & 7.5                                                & 7.5                                                & 7.5  & 7.5                                                  & \textbf{15} \\
		utbm  & 10                                                 & 10                                                 & 10   & 10                                                   & \textbf{20} \\
		ulhk  & 10                                                 & 10                                                 & 10   & 10                                                   & \textbf{20} \\
		kaist & 10                                                 & 10                                                 & 10   & 10                                                   & \textbf{20} \\ \hline
	\end{tabular}
\end{center}
\end{table}

Results in Table \ref{table4} demonstrate that our sweep reconstruction method can provide higher-frequency sweep than existing state-of-the-arts. Limited by the frequency of LiDAR scan (e.g., 10\,Hz), the frequency of iterated state update can only be performed at a maximum frequency of 10\,Hz. Even if the power of computing resources (e.g., CPU) is assumed to be infinite, the iterated state update cannot be performed at higher than 10\,Hz. However, with the assistance of our sweep reconstruction, the frequency of LiDAR sweep can be increased to an arbitrary value theoretically. In this work, we increase the frequency of LiDAR sweep from 10\,Hz to 20\,Hz. Benefit from the very low computational overhead of iEKF framework, our system can still run in real time with 20\,Hz reconstructed sweeps.

\subsection{Ablation Study of Sweep Reconstruction}
\label{Ablation Study of Sweep Reconstruction}

We examine the impact of sweep reconstruction on pose estimation accuracy by comparing the ATE result with vs. without sweep reconstruction. Without using the proposed sweep reconstruction, the system takes 10\,Hz raw input to perform state update at 10\,Hz. In addition, all other configuration parameters are unchanged.

\begin{table}[]
\begin{center}
	\caption{Ablation Study of Sweep Reconstruction on RMSE of ATE (Unit: m)}
	\label{table5}
	\begin{tabular}{c|c|c}
		\hline
		& Ours w/o sweep reconstruction & Ours           \\ \hline
		$nclt\_1$  & 1.40                          & \textbf{1.34}  \\
		$nclt\_2$  & 1.94                          & \textbf{1.80}  \\
		$nclt\_3$  & 5.25                          & \textbf{2.37}  \\
		$nclt\_4$  & $\times$                             & \textbf{1.91}  \\
		$nclt\_5$  & $\times$                             & \textbf{1.62}  \\
		$nclt\_6$  & 2.24                          & \textbf{2.10}  \\
		$nclt\_7$  & 2.23                          & \textbf{2.13}  \\
		$nclt\_8$  & $\times$                             & \textbf{2.70}  \\
		$nclt\_9$  & 4.00                          & \textbf{2.11}  \\
		$nclt\_10$ & 1.97                          & \textbf{1.67}  \\
		$nclt\_11$ & 1.85                          & \textbf{1.61}  \\
		$utbm\_1$  & 10.74                         & \textbf{7.70}  \\
		$utbm\_2$  & 16.98                         & \textbf{16.28} \\
		$utbm\_3$  & 9.94                          & \textbf{8.42}  \\
		$utbm\_4$  & 11.35                         & \textbf{11.12} \\
		$utbm\_5$  & 10.04                         & \textbf{9.14}  \\
		$ulhk\_1$  & \textbf{1.02}                 & \textbf{1.02}  \\
		$ulhk\_2$  & 3.36                          & \textbf{3.21}  \\
		$kaist\_1$ & 1.14                          & \textbf{1.10}  \\
		$kaist\_2$ & 0.95                          & \textbf{0.92}  \\
		$kaist\_3$ & 1.56                          & \textbf{1.36}  \\ \hline
	\end{tabular}
\end{center}
\end{table}

Results in Table \ref{table5} demonstrate that increasing the frequency of state update by increasing sweep frequency can improve the accuracy and robustness of iEKF framework.

\subsection{Ablation Study of Distortion Correction}
\label{Ablation Study of Distortion Correction}

\begin{figure}
	\begin{center}
		\includegraphics[scale=0.5]{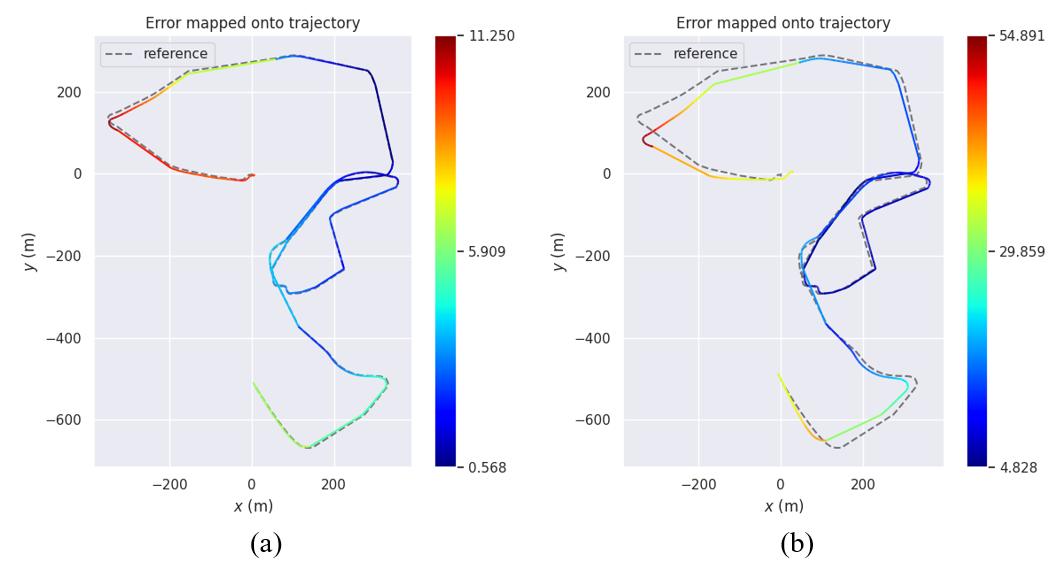}
		\caption{The comparison of partial trajectory of $utbm\_1$ with (a) distortion correction for each segment and (b) distortion correction for the reconstructed sweep directly.}
		\label{fig4}
	\end{center}
\end{figure}

As illustrated in Sec. \ref{Distortion Correction}, the proposed distortion correction for each segment can prevent inaccurate trajectory caused by multiple inconsistent distortion correction to a particular point. Therefore, we evaluate the effectiveness of this distortion correction method in this section. As shown in Fig. \ref{fig4}, (a) is the partial trajectory of exemplar sequence $utbm\_1$ with distortion correction for each segment, (b) is the partial trajectory of exemplar sequence $utbm\_1$ with distortion correction for the reconstructed sweep directly. Only partial trajectories are compared because drift occurs halfway through case (b). The comparison result illustrates that the proposed distortion correction for each segment can better undistort LiDAR points, and in turn obtain accurate trajectory.

\subsection{Time Consumption}
\label{Time Consumption}

\begin{table}[]
\begin{center}
	\caption{Time Consumption Per Reconstructed Sweep (Unit: ms)}
	\label{table6}
	\begin{tabular}{c|cccc|c|c}
		\hline
		& (1)  & (2)  & (3)   & (4)   & Total & \begin{tabular}[c]{@{}c@{}}Maximum\\ Available\\ Time\end{tabular} \\ \hline
		$nclt\_1$  & 1.86 & 4.35 & 11.24 & 10.38 & 28.32 & 65                                                                 \\
		$nclt\_2$  & 1.81 & 4.22 & 12.07 & 9.72  & 28.35 & 65                                                                 \\
		$nclt\_3$  & 1.76 & 3.82 & 11.76 & 9.77  & 27.62 & 65                                                                 \\
		$nclt\_4$  & 1.94 & 3.96 & 11.46 & 10.83 & 28.78 & 65                                                                 \\
		$nclt\_5$  & 1.87 & 4.39 & 13.66 & 10.28 & 30.77 & 65                                                                 \\
		$nclt\_6$  & 2.07 & 4.28 & 13.00 & 9.93  & 29.85 & 65                                                                 \\
		$nclt\_7$  & 2.02 & 4.37 & 12.34 & 7.97  & 27.37 & 65                                                                 \\
		$nclt\_8$  & 2.09 & 4.59 & 13.78 & 5.62  & 26.79 & 65                                                                 \\
		$nclt\_9$  & 2.02 & 4.26 & 12.45 & 10.51 & 29.90 & 65                                                                 \\
		$nclt\_10$ & 2.10 & 4.15 & 11.94 & 10.37 & 29.10 & 65                                                                 \\
		$nclt\_11$ & 1.85 & 3.89 & 11.48 & 9.29  & 27.07 & 65                                                                 \\
		$utbm\_1$  & 3.73 & 3.71 & 8.44  & 4.73  & 21.10 & 50                                                                 \\
		$utbm\_2$  & 3.99 & 3.86 & 9.32  & 4.76  & 22.45 & 50                                                                 \\
		$utbm\_3$  & 3.87 & 3.60 & 9.68  & 4.96  & 22.71 & 50                                                                 \\
		$utbm\_4$  & 3.83 & 3.61 & 8.24  & 5.06  & 21.25 & 50                                                                 \\
		$utbm\_5$  & 3.80 & 2.85 & 7.96  & 5.53  & 20.64 & 50                                                                 \\
		$ulhk\_1$  & 7.92 & 3.53 & 7.28  & 1.10  & 20.34 & 50                                                                 \\
		$ulhk\_2$  & 9.48 & 3.07 & 6.04  & 1.85  & 21.02 & 50                                                                 \\
		$kaist\_1$ & 1.94 & 3.79 & 10.29 & 5.56  & 22.16 & 50                                                                 \\
		$kaist\_2$ & 1.50 & 3.84 & 11.03 & 4.03  & 20.97 & 50                                                                 \\
		$kaist\_3$ & 1.13 & 3.21 & 11.11 & 5.98  & 21.94 & 50                                                                 \\ \hline
	\end{tabular}
\end{center}
\end{table}

We evaluate the runtime breakdown (unit: ms) of our system for all sequences. In general, the most time-consuming modules are (1) sweep segmentation, (2) sweep reconstruction with distortion correction for each segment, (3) iEKF based state estimation and (4) map update. Therefore, for each sequence, we test the time cost of above four modules, and the total time for handling a sweep.

Results in Table \ref{table6} show that our SR-LIO takes 20$\sim$30\,ms to handle a sweep, while the time interval of two consecutive reconstructed sweep is 65\,ms on $nclt$ dataset, and 50\,ms on $utbm$, $ulhk$, $kaist$ dataset. That means our system can not only run in real time, but also save 20$\sim$45\,ms per reconstructed sweep.

\section{Conclusion}
\label{Conclusion}

This paper utilizes the previous proposed sweep reconstruction method \cite{yuan2023sdv} to increase the frequency of sweeps, and utilizes the higher frequent reconstructed sweeps to perform state update for iEKF based LIO framework (i.e., SR-LIO). For LIO systems, this method can not only increase the frequency of LiDAR sweep, but also reduce the accumulative error of predicted state by reducing the time interval of IMU measurements integration, and in turn achieve more accurate and robust state estimation results. In addition, for reconstructed sweeps with common data, we propose to perform distortion correction for each segment but not an entire reconstructed sweep, to better ensure the accuracy of estimated trajectory.

The proposed SR-LIO achieves the state-of-the-art accuracy on four public datasets. Meanwhile, we demonstrate the effectiveness of sweep reconstruction to improve the output frequency and accuracy of iEKF based LIO systems.

\bibliographystyle{IEEEtrans}
\bibliography{IEEEabrv,IEEEExample}

\begin{thebibliography}{10}
\providecommand{\url}[1]{#1}
\csname url@samestyle\endcsname
\providecommand{\newblock}{\relax}
\providecommand{\bibinfo}[2]{#2}
\providecommand{\BIBentrySTDinterwordspacing}{\spaceskip=0pt\relax}
\providecommand{\BIBentryALTinterwordstretchfactor}{4}
\providecommand{\BIBentryALTinterwordspacing}{\spaceskip=\fontdimen2\font plus
\BIBentryALTinterwordstretchfactor\fontdimen3\font minus
  \fontdimen4\font\relax}
\providecommand{\BIBforeignlanguage}[2]{{%
\expandafter\ifx\csname l@#1\endcsname\relax
\typeout{** WARNING: IEEEtranS.bst: No hyphenation pattern has been}%
\typeout{** loaded for the language `#1'. Using the pattern for}%
\typeout{** the default language instead.}%
\else
\language=\csname l@#1\endcsname
\fi
#2}}
\providecommand{\BIBdecl}{\relax}
\BIBdecl

\bibitem{Agarwal_Ceres_Solver_2022}
\BIBentryALTinterwordspacing
S.~Agarwal, K.~Mierle, and T.~C.~S. Team, ``{Ceres Solver},'' 3 2022. [Online].
  Available: \url{https://github.com/ceres-solver/ceres-solver}
\BIBentrySTDinterwordspacing

\bibitem{behley2018efficient}
J.~Behley and C.~Stachniss, ``Efficient surfel-based slam using 3d laser range
  data in urban environments.'' in \emph{Robotics: Science and Systems}, vol.
  2018, 2018, p.~59.

\bibitem{blanco2014nanoflann}
J.~L. Blanco and P.~K. Rai, ``nanoflann: a {C}++ header-only fork of {FLANN}, a
  library for nearest neighbor ({NN}) with kd-trees,''
  \url{https://github.com/jlblancoc/nanoflann}, 2014.

\bibitem{cai2021ikd}
Y.~Cai, W.~Xu, and F.~Zhang, ``ikd-tree: An incremental kd tree for robotic
  applications,'' \emph{arXiv preprint arXiv:2102.10808}, 2021.

\bibitem{carlevaris2016university}
N.~Carlevaris-Bianco, A.~K. Ushani, and R.~M. Eustice, ``University of michigan
  north campus long-term vision and lidar dataset,'' \emph{The International
  Journal of Robotics Research}, vol.~35, no.~9, pp. 1023--1035, 2016.

\bibitem{chen2023direct}
K.~Chen, R.~Nemiroff, and B.~T. Lopez, ``Direct lidar-inertial odometry:
  Lightweight lio with continuous-time motion correction,'' in \emph{2023 IEEE
  International Conference on Robotics and Automation (ICRA)}.\hskip 1em plus
  0.5em minus 0.4em\relax IEEE, 2023, pp. 3983--3989.

\bibitem{dellenbach2022ct}
P.~Dellenbach, J.-E. Deschaud, B.~Jacquet, and F.~Goulette, ``Ct-icp: Real-time
  elastic lidar odometry with loop closure,'' in \emph{2022 International
  Conference on Robotics and Automation (ICRA)}.\hskip 1em plus 0.5em minus
  0.4em\relax IEEE, 2022, pp. 5580--5586.

\bibitem{deschaud2018imls}
J.-E. Deschaud, ``Imls-slam: Scan-to-model matching based on 3d data,'' in
  \emph{2018 IEEE International Conference on Robotics and Automation
  (ICRA)}.\hskip 1em plus 0.5em minus 0.4em\relax IEEE, 2018, pp. 2480--2485.

\bibitem{geneva2020openvins}
P.~Geneva, K.~Eckenhoff, W.~Lee, Y.~Yang, and G.~Huang, ``Openvins: A research
  platform for visual-inertial estimation,'' in \emph{2020 IEEE International
  Conference on Robotics and Automation (ICRA)}.\hskip 1em plus 0.5em minus
  0.4em\relax IEEE, 2020, pp. 4666--4672.

\bibitem{jeong2019complex}
J.~Jeong, Y.~Cho, Y.-S. Shin, H.~Roh, and A.~Kim, ``Complex urban dataset with
  multi-level sensors from highly diverse urban environments,'' \emph{The
  International Journal of Robotics Research}, vol.~38, no.~6, pp. 642--657,
  2019.

\bibitem{li2021towards}
K.~Li, M.~Li, and U.~D. Hanebeck, ``Towards high-performance
  solid-state-lidar-inertial odometry and mapping,'' \emph{IEEE Robotics and
  Automation Letters}, vol.~6, no.~3, pp. 5167--5174, 2021.

\bibitem{qin2020lins}
C.~Qin, H.~Ye, C.~E. Pranata, J.~Han, S.~Zhang, and M.~Liu, ``Lins: A
  lidar-inertial state estimator for robust and efficient navigation,'' in
  \emph{2020 IEEE international conference on robotics and automation
  (ICRA)}.\hskip 1em plus 0.5em minus 0.4em\relax IEEE, 2020, pp. 8899--8906.

\bibitem{qin2018vins}
T.~Qin, P.~Li, and S.~Shen, ``Vins-mono: A robust and versatile monocular
  visual-inertial state estimator,'' \emph{IEEE Transactions on Robotics},
  vol.~34, no.~4, pp. 1004--1020, 2018.

\bibitem{shan2018lego}
T.~Shan and B.~Englot, ``Lego-loam: Lightweight and ground-optimized lidar
  odometry and mapping on variable terrain,'' in \emph{2018 IEEE/RSJ
  International Conference on Intelligent Robots and Systems (IROS)}.\hskip 1em
  plus 0.5em minus 0.4em\relax IEEE, 2018, pp. 4758--4765.

\bibitem{shan2020lio}
T.~Shan, B.~Englot, D.~Meyers, W.~Wang, C.~Ratti, and D.~Rus, ``Lio-sam:
  Tightly-coupled lidar inertial odometry via smoothing and mapping,'' in
  \emph{2020 IEEE/RSJ international conference on intelligent robots and
  systems (IROS)}.\hskip 1em plus 0.5em minus 0.4em\relax IEEE, 2020, pp.
  5135--5142.

\bibitem{sorenson1966kalman}
H.~W. Sorenson, ``Kalman filtering techniques,'' in \emph{Advances in control
  systems}.\hskip 1em plus 0.5em minus 0.4em\relax Elsevier, 1966, vol.~3, pp.
  219--292.

\bibitem{wang2021f}
H.~Wang, C.~Wang, C.-L. Chen, and L.~Xie, ``F-loam: Fast lidar odometry and
  mapping,'' in \emph{2021 IEEE/RSJ International Conference on Intelligent
  Robots and Systems (IROS)}.\hskip 1em plus 0.5em minus 0.4em\relax IEEE,
  2021, pp. 4390--4396.

\bibitem{wen2020urbanloco}
W.~Wen, Y.~Zhou, G.~Zhang, S.~Fahandezh-Saadi, X.~Bai, W.~Zhan, M.~Tomizuka,
  and L.-T. Hsu, ``Urbanloco: A full sensor suite dataset for mapping and
  localization in urban scenes,'' in \emph{2020 IEEE international conference
  on robotics and automation (ICRA)}.\hskip 1em plus 0.5em minus 0.4em\relax
  IEEE, 2020, pp. 2310--2316.

\bibitem{xu2022fast}
W.~Xu, Y.~Cai, D.~He, J.~Lin, and F.~Zhang, ``Fast-lio2: Fast direct
  lidar-inertial odometry,'' \emph{IEEE Transactions on Robotics}, vol.~38,
  no.~4, pp. 2053--2073, 2022.

\bibitem{xu2021fast}
W.~Xu and F.~Zhang, ``Fast-lio: A fast, robust lidar-inertial odometry package
  by tightly-coupled iterated kalman filter,'' \emph{IEEE Robotics and
  Automation Letters}, vol.~6, no.~2, pp. 3317--3324, 2021.

\bibitem{yan2020eu}
Z.~Yan, L.~Sun, T.~Krajn{\'\i}k, and Y.~Ruichek, ``Eu long-term dataset with
  multiple sensors for autonomous driving,'' in \emph{2020 IEEE/RSJ
  International Conference on Intelligent Robots and Systems (IROS)}.\hskip 1em
  plus 0.5em minus 0.4em\relax IEEE, 2020, pp. 10\,697--10\,704.

\bibitem{ye2019tightly}
H.~Ye, Y.~Chen, and M.~Liu, ``Tightly coupled 3d lidar inertial odometry and
  mapping,'' in \emph{2019 International Conference on Robotics and Automation
  (ICRA)}.\hskip 1em plus 0.5em minus 0.4em\relax IEEE, 2019, pp. 3144--3150.

\bibitem{yuan2023sdv}
Z.~Yuan, Q.~Wang, K.~Cheng, T.~Hao, and X.~Yang, ``Sdv-loam: Semi-direct
  visual-lidar odometry and mapping,'' \emph{IEEE Transactions on Pattern
  Analysis and Machine Intelligence}, 2023.

\bibitem{zhang2014loam}
J.~Zhang and S.~Singh, ``Loam: Lidar odometry and mapping in real-time.'' in
  \emph{Robotics: Science and systems}, vol.~2, no.~9.\hskip 1em plus 0.5em
  minus 0.4em\relax Berkeley, CA, 2014, pp. 1--9.

\bibitem{zhang2017low}
------, ``Low-drift and real-time lidar odometry and mapping,''
  \emph{Autonomous Robots}, vol.~41, pp. 401--416, 2017.

\end{thebibliography}

\end{document}